%% file: main.tex
\documentclass[10pt,conference]{IEEEtran}

\usepackage{multirow} %
\usepackage{makecell}
\usepackage{arydshln}
\usepackage[normalem]{ulem}
\usepackage{balance}

\usepackage[sort,compress]{cite}

\ifCLASSINFOpdf
   \usepackage[pdftex]{graphicx}
   \graphicspath{{figs/}}
   \DeclareGraphicsExtensions{.pdf,.jpeg,.png}
\else
   \usepackage[dvips]{graphicx}
   \graphicspath{{../figs/}}
   \DeclareGraphicsExtensions{.eps}
\fi

\usepackage[cmex10]{amsmath}
\usepackage{textcomp} %
\usepackage{gensymb} %
\usepackage{amsthm}

\usepackage{algorithmic}

\usepackage{array}

\ifCLASSOPTIONcompsoc
  \usepackage[caption=false,font=normalsize,labelfont=sf,textfont=sf,farskip=0pt]{subfig}
\else
  \usepackage[caption=false,font=footnotesize,farskip=0pt]{subfig}
\fi

\usepackage[acronym]{glossaries}
\usepackage{xspace}
\usepackage[hyphens]{url}
\usepackage{booktabs}
\usepackage[dvipsnames]{xcolor}%
\usepackage[utf8]{inputenc}
\usepackage[T1]{fontenc}

\newif\ifieee
 \ieeefalse

\ifieee
\else
\usepackage[colorlinks=true,allcolors=black]{hyperref} %
\fi

\usepackage[capitalise]{cleveref} %

\newacronymstyle{long-short-br}
{%
  \GlsUseAcrEntryDispStyle{long-short}%
}%
{%
  \GlsUseAcrStyleDefs{long-short}%
}
\setacronymstyle{long-short-br}

\usepackage{transparent}
\usepackage{tikz}
\newcommand\copyrighttext{%
  \scriptsize Accepted at SIBGRAPI 2024. The final published version is available on IEEE Xplore (DOI: \href{https://doi.org/10.1109/SIBGRAPI62404.2024.10716307}{\textcolor{blue}{10.1109/SIBGRAPI62404.2024.10716307}}).}
\newcommand\copyrightnotice{%
\begin{tikzpicture}[remember picture,overlay]
\node[anchor=south,yshift=30pt,xshift=0pt] at (current page.south) {\fbox{\transparent{0.85}\parbox{\dimexpr0.77\textwidth-\fboxsep-\fboxrule\relax}{\copyrighttext}}};
\end{tikzpicture}%
}

\newif\iffinal
\finaltrue
\newcommand{\cmtid}{54}

\iffinal
\else
\usepackage[switch]{lineno}
\fi

\newcommand*{\RL}[2][]{\textcolor{Rhodamine}{[\textbf{\ifthenelse{\equal{#1}{}}{RL}{RL(#1)}}: #2]}}
\newcommand*{\DM}[2][]{\textcolor{orange}{[\textbf{\ifthenelse{\equal{#1}{}}{DM}{DM(#1)}}: #2]}}
\newcommand*{\EN}[2][]{\textcolor{orange}{[\textbf{\ifthenelse{\equal{#1}{}}{EN}{EN(#1)}}: #2]}}
\newcommand*{\ES}[2][]{\textcolor{orange}{[\textbf{\ifthenelse{\equal{#1}{}}{ES}{ES(#1)}}: #2]}}
\newcommand*{\GL}[2][]{\textcolor{orange}{[\textbf{\ifthenelse{\equal{#1}{}}{ES}{ES(#1)}}: #2]}}

\newcounter{fncounter}
\newcommand\customfootnote[1]{\stepcounter{fncounter}\footnote{\hspace{0.25mm}#1}}

\begin{document}

\input{0-variables}

\title{Toward Enhancing Vehicle Color Recognition in Adverse Conditions: A Dataset and Benchmark}

\iffinal

\author{
\IEEEauthorblockN{Gabriel E. Lima\IEEEauthorrefmark{1}, Rayson Laroca\IEEEauthorrefmark{2}$^,$\IEEEauthorrefmark{1}, Eduardo Santos\IEEEauthorrefmark{3}$^,$\IEEEauthorrefmark{1}, Eduil Nascimento~Jr.\IEEEauthorrefmark{3}, and David Menotti\IEEEauthorrefmark{1}}
\IEEEauthorblockA{
    \IEEEauthorrefmark{1}Department of Informatics, Federal University of Paran\'{a}, Curitiba, Brazil \\
    \IEEEauthorrefmark{2}Postgraduate
    Program in Informatics, Pontifical Catholic University of Paran\'a, Curitiba, Brazil \\
    \IEEEauthorrefmark{3}Department of Technological Development and Quality, Paran\'{a} Military Police, Curitiba, Brazil \\
    \resizebox{0.925\linewidth}{!}{
        \hspace{-0.75mm}\IEEEauthorrefmark{1}\hspace{-0.35mm}\tt{\small{\{gelima,menotti\}}@inf.ufpr.br} \quad \IEEEauthorrefmark{2}{\tt\small rayson@ppgia.pucpr.br} \quad \IEEEauthorrefmark{3}\hspace{-0.2mm}{\tt\small \{{ed.santos,eduiljunior\}}@pm.pr.gov.br}}
    }
}

\fi

\maketitle

\ifieee
    {\let\thefootnote\relax\footnote{\\979-8-3503-7603-6/24/\$31.00
    \textcopyright2024 IEEE}}
\else
    \copyrightnotice
\fi

\input{0-acronyms}

\ifieee
\vspace{-3.575mm}
\else
\vspace{-3.575mm}
\fi
\begin{abstract}

Vehicle information recognition is crucial in various practical domains, particularly in criminal investigations.
\gls*{vcr} has garnered significant research interest because color is a visually distinguishable attribute of vehicles and is less affected by partial occlusion and changes in viewpoint.
Despite the success of existing methods for this task, the relatively low complexity of the datasets used in the literature has been largely overlooked.
This research addresses this gap by compiling a new dataset representing a more challenging \gls*{vcr} scenario.
The images --~sourced from six license plate recognition datasets~-- are categorized into eleven colors, and their annotations were validated using official vehicle registration information.
We evaluate the performance of four deep learning models on a widely adopted dataset and our proposed dataset to establish a benchmark.
The results demonstrate that our dataset poses greater difficulty for the tested models and highlights scenarios that require further exploration in VCR.
Remarkably, nighttime scenes account for a significant portion of the errors made by the best-performing model.
This research provides a foundation for future studies on \gls*{vcr}, while also offering valuable insights for the field of fine-grained vehicle~classification.
\end{abstract}

\IEEEpeerreviewmaketitle

\input{1-introduction}

\input{2-related_work}
\input{3-dataset}
\input{4-experiments}
\input{5-conclusions}
\input{0-acknowledgments}

\balance

\bibliographystyle{IEEEtran}
\bibliography{bibtex}

\end{document}

%% file: 0-variables.tex
\iffinal
\newcommand{\dataset}{UFPR-VCR\xspace}
\newcommand{\urlDataset}{\url{https://github.com/lima001/ufpr-vcr-dataset}}

\fi

%% file: 0-acronyms.tex
\newacronym{alpr}{ALPR}{Automatic License Plate Recognition}
\newacronym{capes}{CAPES}{Coordination for the Improvement of Higher Education Personnel}
\newacronym{cnn}{CNN}{Convolutional Neural Network}
\newacronym{cnpq}{CNPq}{National Council for Scientific and Technological Development}
\newacronym{dl}{DL}{Deep Learning}
\newacronym{senatran}{SENATRAN}{Brazil's National Traffic Secretariat}
\newacronym{fpn}{FPN}{Feature Pyramid Networks}
\newacronym{its}{ITS}{Intelligent Transportation Systems}
\newacronym{mcff}{MCFF-CNN}{Multiscale Comprehensive Feature Fusion Convolutional Neural Network}
\newacronym{mlp}{MLP}{Multi-Layer Perceptron}
\newacronym{pca}{PCA}{Principal Component Analysis}
\newacronym{roi}{ROI}{Region of Interest}
\newacronym{spp}{SPP}{Spatial Pyramid Pooling}
\newacronym{svm}{SVM}{Support Vector Machine}
\newacronym{vcr}{VCR}{Vehicle Color Recognition}
\newacronym{vit}{ViT}{Vision Transformer}

\iffinal
    \newacronym{dataset}{\dataset}{UFPR Vehicle Color Recognition}
\fi

%% file: 1-introduction.tex
\section{Introduction}

\glsresetall

Over the past two decades, there has been significant interest in extracting vehicle information from images taken by surveillance cameras. 
This technology plays a crucial role in various practical domains~\cite{chen2014vehicle,zhang2018vehicle,fu2018mcff,hu2023vehicle}, especially in criminal investigations.
Within this context, \gls{vcr} holds significant importance.
Color covers a substantial portion of the vehicle's surface, making it less prone to partial occlusion and less affected by changes in viewpoint~\cite{hu2015vehicle,wang2021transformer}.

Previous research on \gls*{vcr} can be broadly categorized into two main groups: handcrafted methods~\cite{baek2007vehicle,son2007convolution,dule2010convenient,chen2014vehicle,hsieh2015vehicle} and deep learning-based approaches~\cite{hu2015vehicle,fu2018mcff,zhang2018vehicle,wang2021transformer,hu2023vehicle}. 
Alongside the approaches, several datasets have been created to tackle the \gls*{vcr} problem. 
Notably, Chen et al.~\cite{chen2014vehicle} introduced the first publicly available dataset, comprising $15{,}601$ vehicles categorized into eight color classes.
\ifieee
    This dataset has become popular~\cite{hu2015vehicle,fu2018mcff,zhang2018vehicle,hu2023vehicle} due to its original challenging nature, characterized by images with lighting variations, haze, and~overexposure.
\else
    This dataset has become a popular choice for subsequent studies~\cite{hu2015vehicle,fu2018mcff,zhang2018vehicle,hu2023vehicle} due to its original challenging nature, characterized by images with lighting variations, haze, and~overexposure.
\fi

\begin{figure}[!t]
\centering
\captionsetup[subfigure]{captionskip=2pt}

\subfloat[Images from the dataset proposed by Chen et al.~\cite{chen2014vehicle}.]
{\includegraphics[width=0.85\columnwidth]{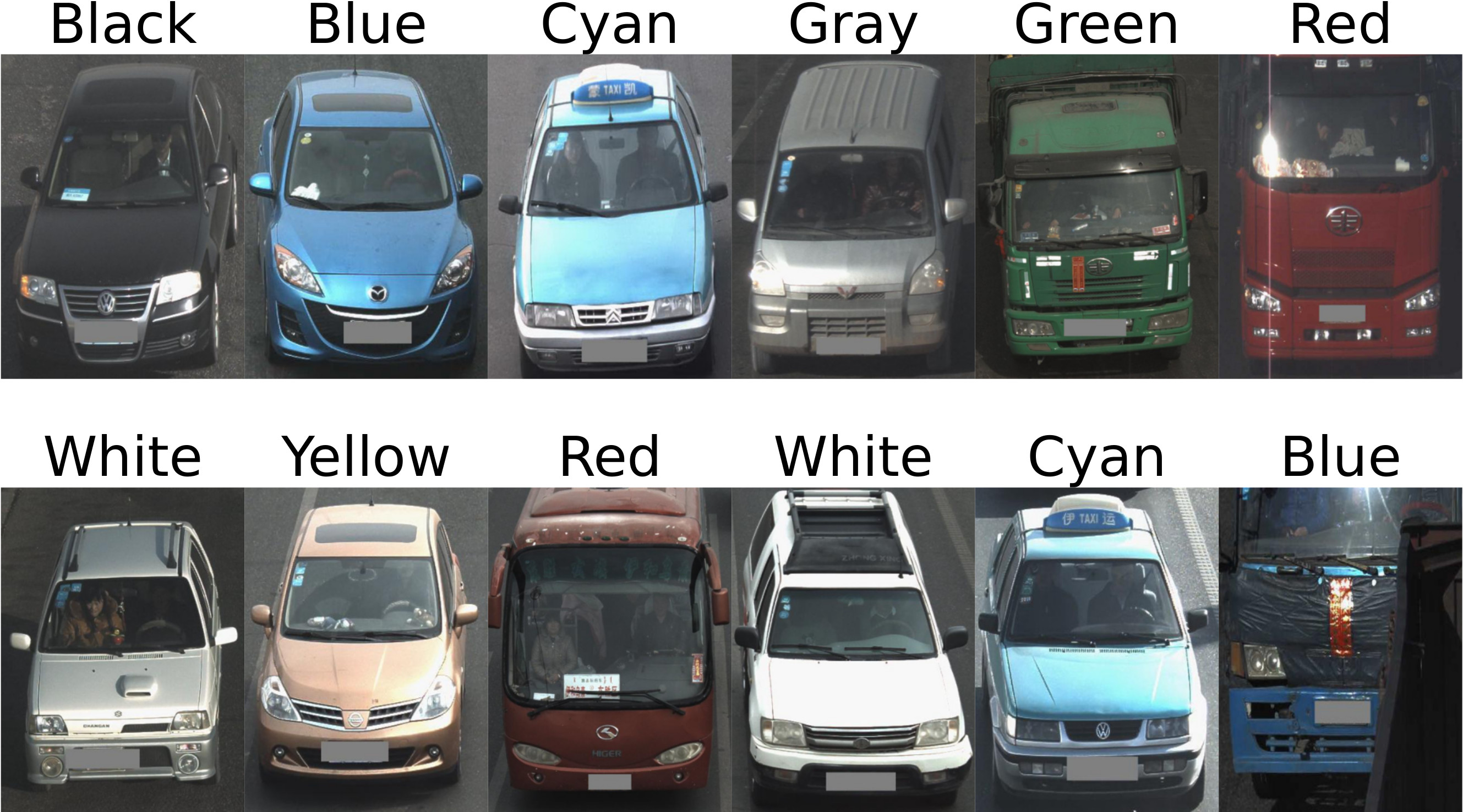}%
\label{fig:grid_chen}}
\vspace{10pt}
\subfloat[Images from the \dataset dataset, proposed in this~work.]{\includegraphics[width=0.875\columnwidth]{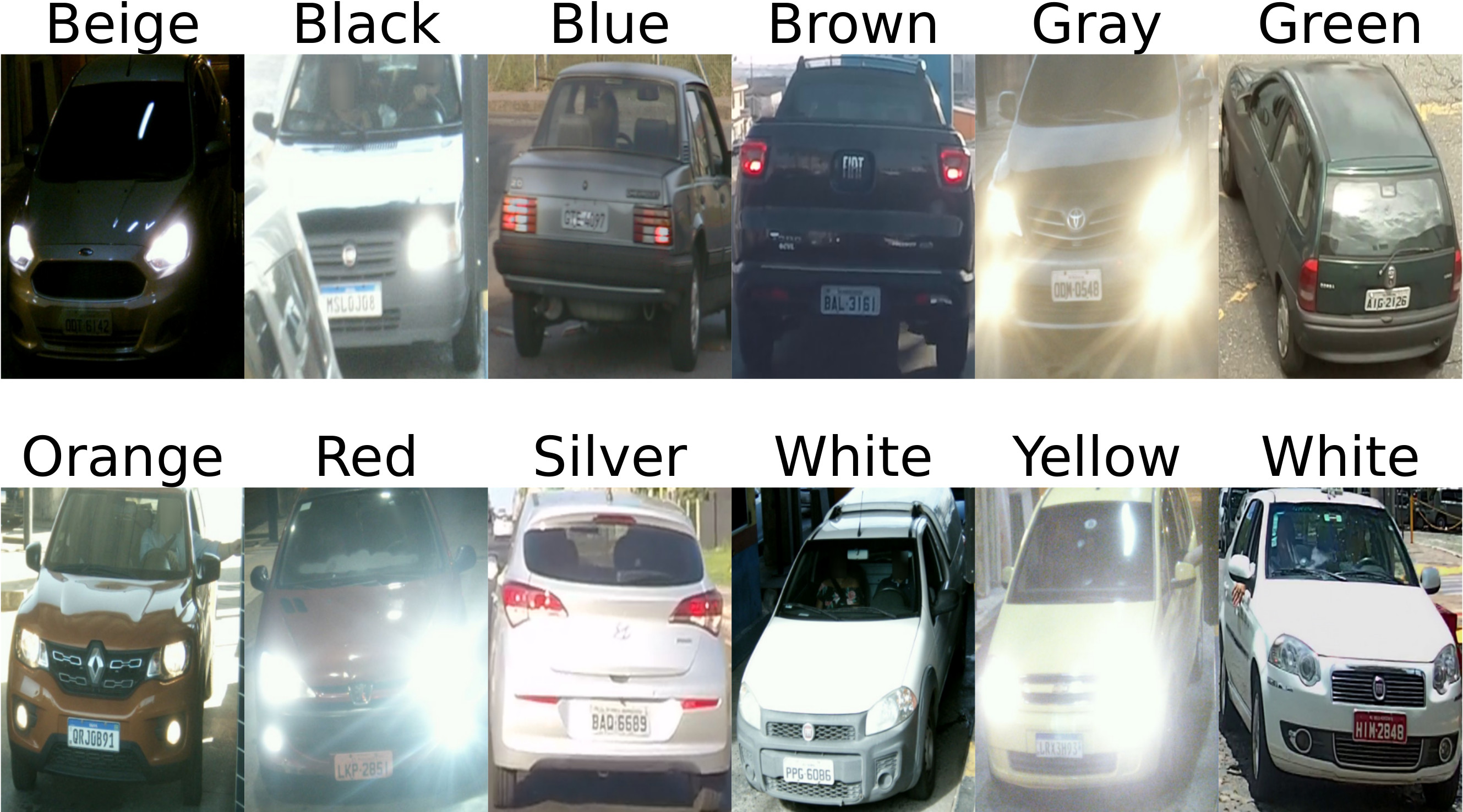}%
\label{fig:grid_ufpr_vc}}
\vspace{0.25mm}
\caption{Examples of images from the datasets proposed in~\cite{chen2014vehicle}~(a) and in this work~(b), with the corresponding vehicle color annotation shown above each image. Observe that images in the proposed dataset~(b) depict significantly more challenging scenes than those in~(a), featuring adverse conditions such as nighttime settings and vehicles from various~viewpoints.}
\label{fig:datasets_grid}
\end{figure}

Despite these conditions, studies have reported satisfactory outcomes, achieving up to $95$\% average accuracy in color recognition across the mentioned dataset and datasets with similar characteristics~\cite{fu2018mcff,zhang2018vehicle,wang2021transformer,hu2023vehicle}.
However, upon analyzing these studies, it becomes evident that the explored datasets lack images depicting highly adverse conditions.
The images primarily feature vehicles captured under adequate lighting conditions and from a consistent viewpoint, with clearly distinguishable colors, as shown in \cref{fig:grid_chen}.
Consequently, the reported satisfactory results may be misleading, as the testing scenarios do not fully comprise the challenges often found in real-world, unconstrained \gls*{vcr}~applications.

\iffinal
    To tackle a more complex \gls*{vcr} scenario, we introduce the \gls*{dataset} dataset\customfootnote{\texttt{\urlDataset}}$^,$\footnote{Access is granted upon request, i.e., interested parties must register by filling out a registration form and agreeing to the dataset’s terms of use.}.
\fi
It comprises $10{,}039$ images featuring various real-world conditions such as frontal and rear views, partially occluded vehicles, diverse and uneven lighting, and nighttime scenes.
The images were sourced from six public datasets collected in Brazil, originally intended for \gls*{alpr}.
The images cover eleven distinct vehicle colors, with annotations for over $90$\% of the vehicles validated using information obtained from the corresponding license plates (see details in \cref{sub_sec:annotations}).

\cref{fig:grid_ufpr_vc} presents sample vehicle images from the proposed dataset, along with their corresponding color annotations.
These images highlight the challenging \gls*{vcr} scenarios considered in this work, demonstrating the difficulty of accurately classifying vehicle colors.
To assess the dataset's challenges and pinpoint areas for improvement in \gls*{vcr}, we conduct a benchmark study using four deep learning-based~models.

The remainder of this work is organized as follows.
\cref{sec:related_work} reviews related works.
\cref{sec:dataset} introduces the \dataset dataset.
\cref{sec:experiments} details the conducted experiments and presents the achieved results.
Finally, \cref{sec:conclusions} summarizes our findings and outlines directions for future~research.

%% file: 2-related_work.tex
\section{Related Work}
\label{sec:related_work}

This section overviews the datasets proposed within the \gls*{vcr} context.
It outlines their key characteristics, describes the methodologies employed in studies using these datasets, and summarizes the results they have~achieved.
 
In 2007, Baek et al.~\cite{baek2007vehicle} proposed a dataset comprising $500$ images from unspecified scenarios, equally distributed across five colors.
They utilized a 2D histogram technique within the HSV color space for feature extraction and employed \glspl{svm} for classification, reporting an average accuracy of $94.9$\%. 
Using this dataset, Son et al.~\cite{son2007convolution} further refined the \gls*{svm} classifier by proposing a convolution kernel, achieving precision and recall rates exceeding~$92$\%.

Three years later, Dule et al.~\cite{dule2010convenient} introduced a dataset containing $1{,}960$ highway images, evenly distributed across seven colors.
Half of the images were pre-processed to include a smoothed hood region of the vehicle, while the remaining images show frontal views.
The highest accuracy ($83.5$\%) was achieved using histogram features from different color spaces, classified with a multilayer~perceptron.

While the datasets mentioned above were once accessible, they are currently unavailable to the best of our knowledge.
In 2014, Chen et al.~\cite{chen2014vehicle} introduced a dataset for \gls*{vcr} consisting of $15{,}601$ frontal images captured by surveillance cameras under challenging conditions, compared to other datasets at the time.
These images are categorized into eight colors: black, blue, cyan, gray, green, red, white, and yellow.
The authors employed an implicit region-of-interest selector integrating spatial information for feature extraction.
By combining this technique with principal component analysis and an \gls{svm} classifier, they attained an average recognition precision of~$92.5$\%.

In subsequent studies using the same dataset, efforts were made to refine the feature extraction method.
Hu et al.~\cite{hu2015vehicle} integrated a \gls*{cnn} with \gls*{spp}, achieving an average precision of $94.6$\%.
Zhang et al.~\cite{zhang2018vehicle} further proposed feature fusion from multiple \gls*{cnn} layers via \gls*{spp}, resulting in an average precision of $95.4$\%.
Fu et al.~\cite{fu2018mcff} introduced the \gls{mcff}, which incorporates residual learning and inception modules, achieving an average precision above~$97$\%.

In 2021, Wang et al.~\cite{wang2021transformer} built a dataset featuring $32{,}220$ vehicles distributed across eleven colors, further subdivided into $75$ subcategories.
The dataset exclusively contains rear-view vehicle images, annotated using a clustering algorithm and prior knowledge of typical vehicle colors.
The researchers achieved an average accuracy of $97.8$\% using a hybrid model that combines \gls*{cnn} and \gls*{vit} models.
Despite our efforts to contact the authors for access to the dataset, we have not yet received a~response.

In 2023, Hu et al.~\cite{hu2023vehicle} introduced the \textit{Vehicle Color-24} dataset, composed of $31{,}232$ vehicles categorized into $24$ color classes.
Before preprocessing, the dataset included $10{,}091$ frontal-view vehicle images suitable for both vehicle detection and color identification tasks.
The authors employed \gls{cnn} and \gls{fpn} modules for multi-scale information fusion, alongside a loss function aimed at addressing the dataset's long-tail distribution.
They reported a mean average precision of $95.0$\%.

Although initially appealing, the Vehicle Color-24 dataset was not considered in this research for two key reasons.
First, we obtained access to the dataset only recently, during the course of this study.
Second, all samples within the dataset underwent preprocessing steps, including haze removal and lighting adjustments, by its creators.
These alterations may have reduced the dataset's diversity and, consequently, its ability to faithfully represent real-world~conditions.

Despite the progress in \gls*{vcr} research with the development of increasingly robust methods, our analysis of existing datasets reveals that they predominantly feature relatively simple scenarios.
These scenarios typically show vehicles in daylight hours, captured from a single perspective, without occlusions or vehicles partially outside the image frame.
While these datasets have enabled methods to achieve high success rates, they do not fully capture the complexities of real-world scenarios.
Therefore, the introduction of the \dataset dataset represents a distinctive contribution to advancing this field.
Furthermore, previous research has focused on maximizing performance on the \gls*{vcr} task.
This research, however, aims to identify the complex scenarios that limit VCR success in unconstrained scenarios and use those insights to drive further development in the~field.

%% file: 3-dataset.tex
\section{The \dataset Dataset}
\label{sec:dataset}

\begin{figure}[!htb]
\centering
\includegraphics[width=0.99\linewidth]{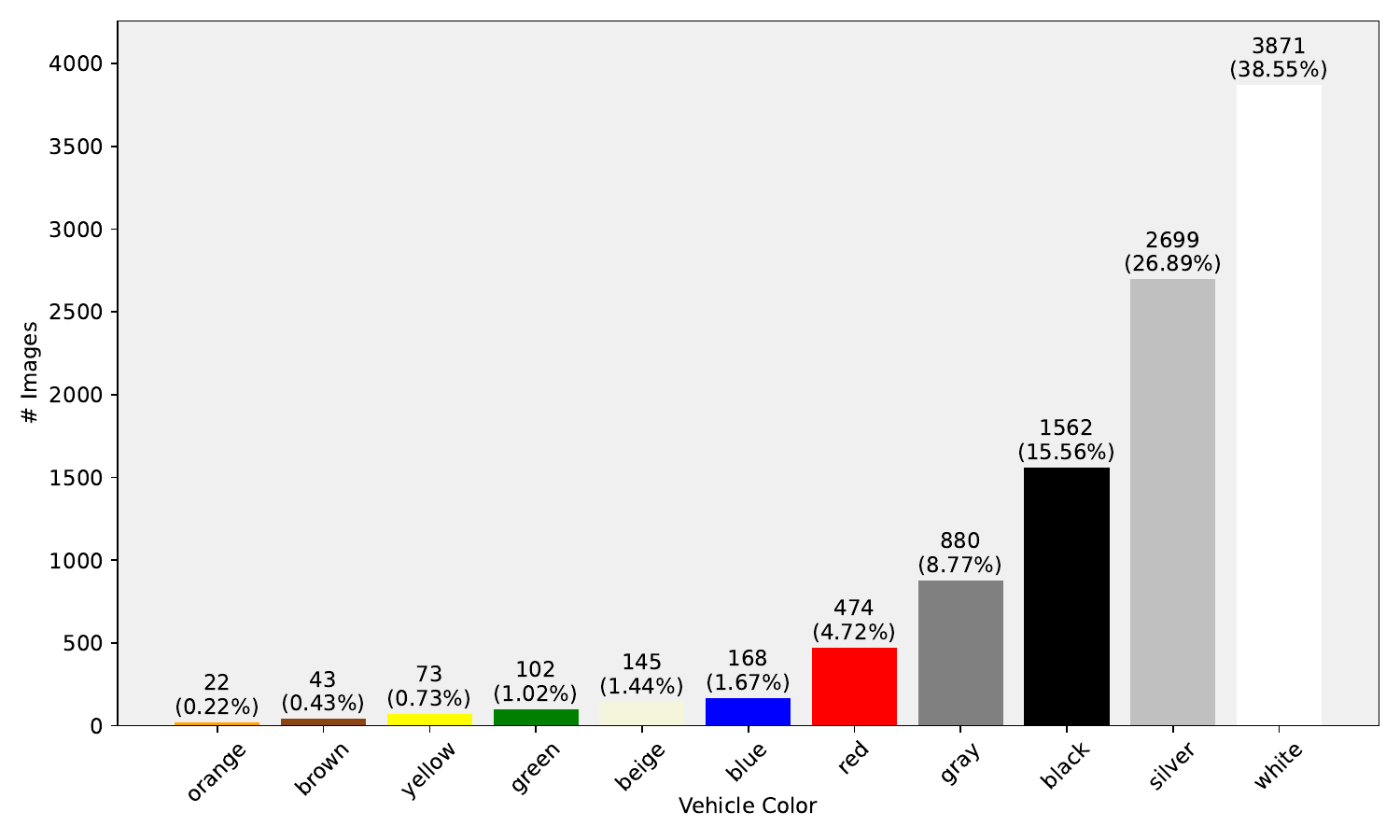}

\vspace{-4mm}

\caption{Distribution of vehicle colors in the \dataset dataset.}
\label{fig:ufpr_vc_img_dist}
\end{figure}

This section details the creation process and main characteristics of the \dataset dataset.
It includes $10{,}039$ images from $9{,}502$ different vehicles, categorized into eleven colors: beige, black, blue, brown, gray, green, orange, red, silver, white, and yellow.
The distribution of images across these colors is highly unbalanced (see \cref{fig:ufpr_vc_img_dist}).
This reflects the real-world scenario, where vehicle colors such as white, black, and shades of gray are more common than others~\cite{axalta2022global,farias2023colorido,harley2023cores}.

The images show vehicles across different categories (i.e., cars, vans, buses, and trucks) captured in diverse environments.
The original images, before vehicle cropping, were sourced from six Brazilian datasets commonly used in \gls*{alpr} research, namely: OpenALPR-BR~\cite{openalpr_br}, RodoSol-ALPR~\cite{laroca2022cross}, SSIG-SegPlate~\cite{goncalves2016benchmark}, UFOP~\cite{mendesjunior2011towards}, UFPR-ALPR~\cite{laroca2018robust}, and Vehicle-Rear~\cite{oliveira2021vehicle}\customfootnote{We received permission from the creators of the explored ALPR datasets to build the \dataset dataset.}.
\cref{tab:alpr_datasets} provides an overview of these~datasets.

\begin{table}[!htb]
    \centering
    \caption{Summary of the \gls*{alpr} datasets used to create~\dataset.}
    \label{tab:alpr_datasets}

    \vspace{-2mm}
    
    \resizebox{0.975\linewidth}{!}{
    \begin{tabular}{lcrcc}
        \toprule
        Dataset &Year &Images\phantom{a} &Resolution &Viewpoint \\
        \midrule
        UFOP~\cite{mendesjunior2011towards} & $2011$ &$\phantom{^\ast}377\phantom{^\ast}$ &$800\times600$ &Frontal/Rear \\
        SSIG-SigPlate~\cite{goncalves2016benchmark} &$2016$ &$\phantom{^\ast}2{,}000\phantom{^\ast}$ & $1920\times1080$ &Frontal \\
        OpenALPR-BR~\cite{openalpr_br} &$2016$ &$\phantom{^\ast}115\phantom{^\ast}$ &Various &Frontal/Rear \\
        UFPR-ALPR~\cite{laroca2018robust} &$2018$ &$\phantom{^\ast}4{,}500\phantom{^\ast}$ & $1920\times1080$ & Frontal/Rear\\
        Vehicle-Rear$^\ast$~\cite{oliveira2021vehicle} & $2021$ &$\phantom{^\ast}445^\ast$ & $1280\times720$ &Rear \\
        RodoSol-ALPR~\cite{laroca2022cross} & $2022$ & $\phantom{^\ast}20{,}000\phantom{^\ast}$ & $1280\times720$ &Frontal/Rear \\ \bottomrule
        \multicolumn{5}{l}{\rule{-3pt}{2.1ex}\scriptsize$^{\ast}$\hspace{0.2mm}We used only the portion of Vehicle-Rear that includes labels for the license plates.} \\
    \end{tabular}
    }
\end{table}

We selected these datasets for three main reasons: (i)~they are widely adopted~\cite{laroca2022cross,laroca2023leveraging}; (ii)~their images depict scenes with diverse lighting and viewpoints; and (iii)~they include license plate-related annotations, enabling the validation of color annotations and therefore minimizing labeling errors.

As expected, the chosen datasets do not share a standard organization scheme.
Hence, preprocessing and image selection procedures were implemented to standardize the images and identify those appropriate for the \gls*{vcr} task.
These procedures are detailed in  \cref{sub_sec:pre_processing} and \cref{sub_sec:image_selection}, respectively.
Lastly, \cref{sub_sec:annotations} describes the labeling~process.

\subsection{Preprocessing}
\label{sub_sec:pre_processing}

\begin{figure}[!hbt]
\centering
\includegraphics[width=0.99\linewidth]{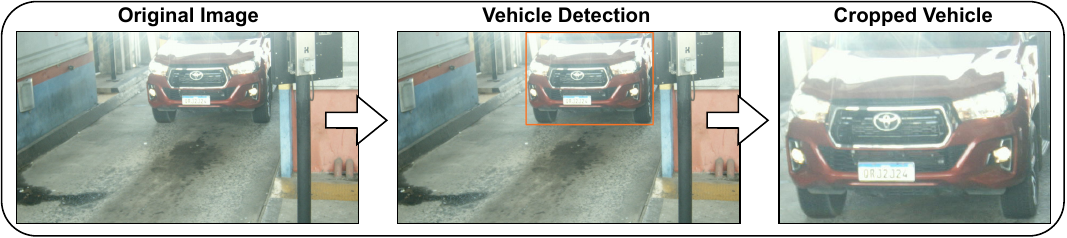}
\vspace{-7mm}
\caption{Illustration depicting the process of extracting vehicle patches from images in the RodoSol-ALPR dataset, which lacks vehicle position~labels.}
\label{fig:preprocessing_pipeline}
\end{figure}

Vehicles were cropped and separated into individual images.
Vehicle bounding box information was readily available in the OpenALPR-BR, SSIG-SegPlate, UFOP, UFPR-ALPR, and Vehicle-Rear datasets.
Due to the lack of this information in the RodoSol-ALPR dataset, we used the well-known YOLOv8 model~\cite{yolov8} for vehicle detection, as depicted in \cref{fig:preprocessing_pipeline}.
This model was chosen for its robust performance and extensive use in both academic research and industry~\cite{yolov8,kim2023high,talaat2023improved}.
Importantly, no noise removal or image enhancement techniques were applied to preserve the original adverse image~conditions.

\subsection{Image selection}
\label{sub_sec:image_selection}

\begin{figure}[!htb]
\centering
\captionsetup[subfigure]{captionskip=1.5pt,font=scriptsize}

\subfloat[Red]{\includegraphics[width=0.3\columnwidth,height=0.25\columnwidth]{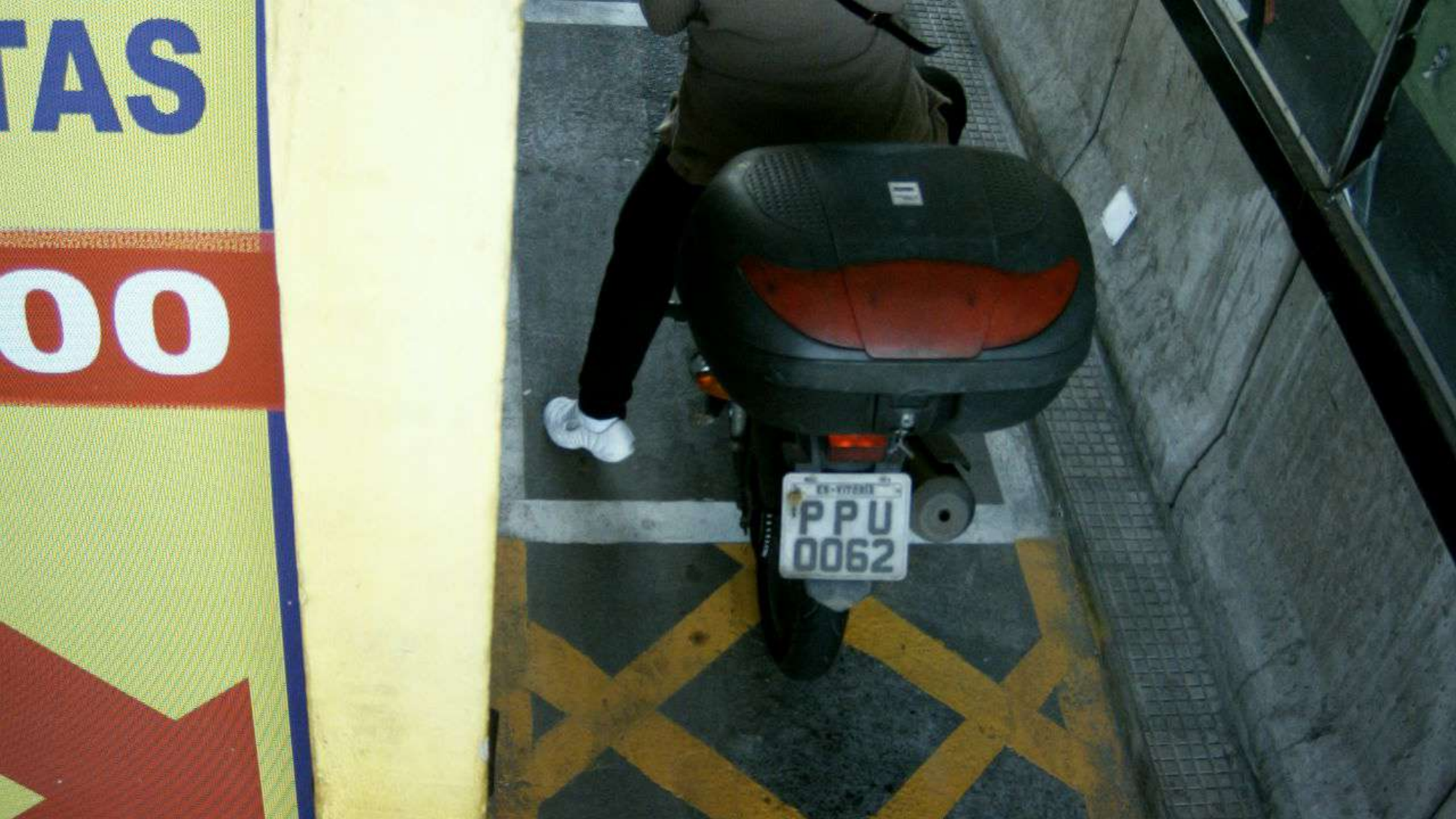}%
\label{fig:motorcycle_1}}
\hspace{0.1mm}
\subfloat[Red]{\includegraphics[width=0.3\columnwidth,height=0.25\columnwidth]{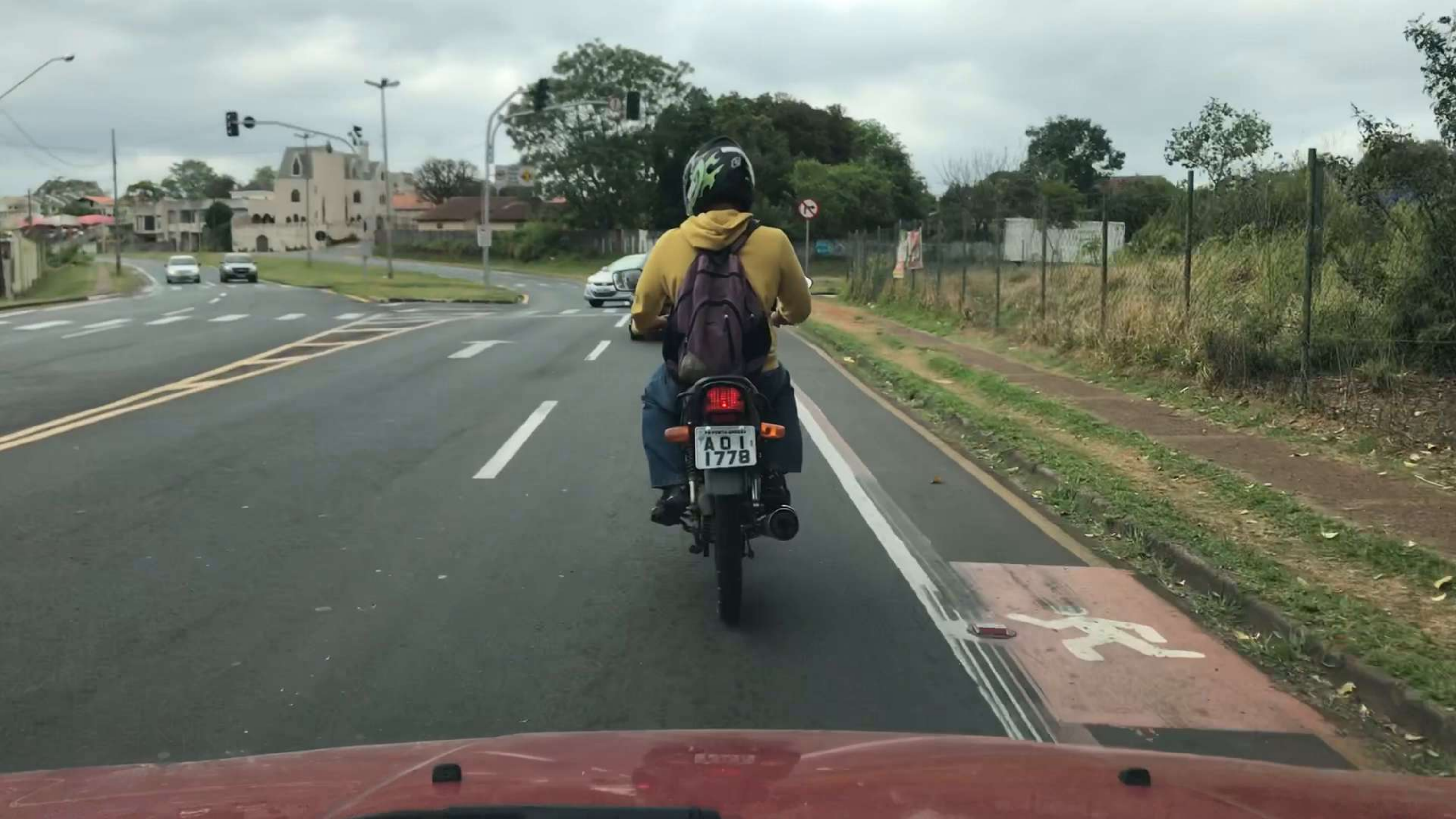}%
\label{fig:motorcycle_3}}
\hspace{0.1mm}
\subfloat[Blue]{\includegraphics[width=0.3\columnwidth,height=0.25\columnwidth]{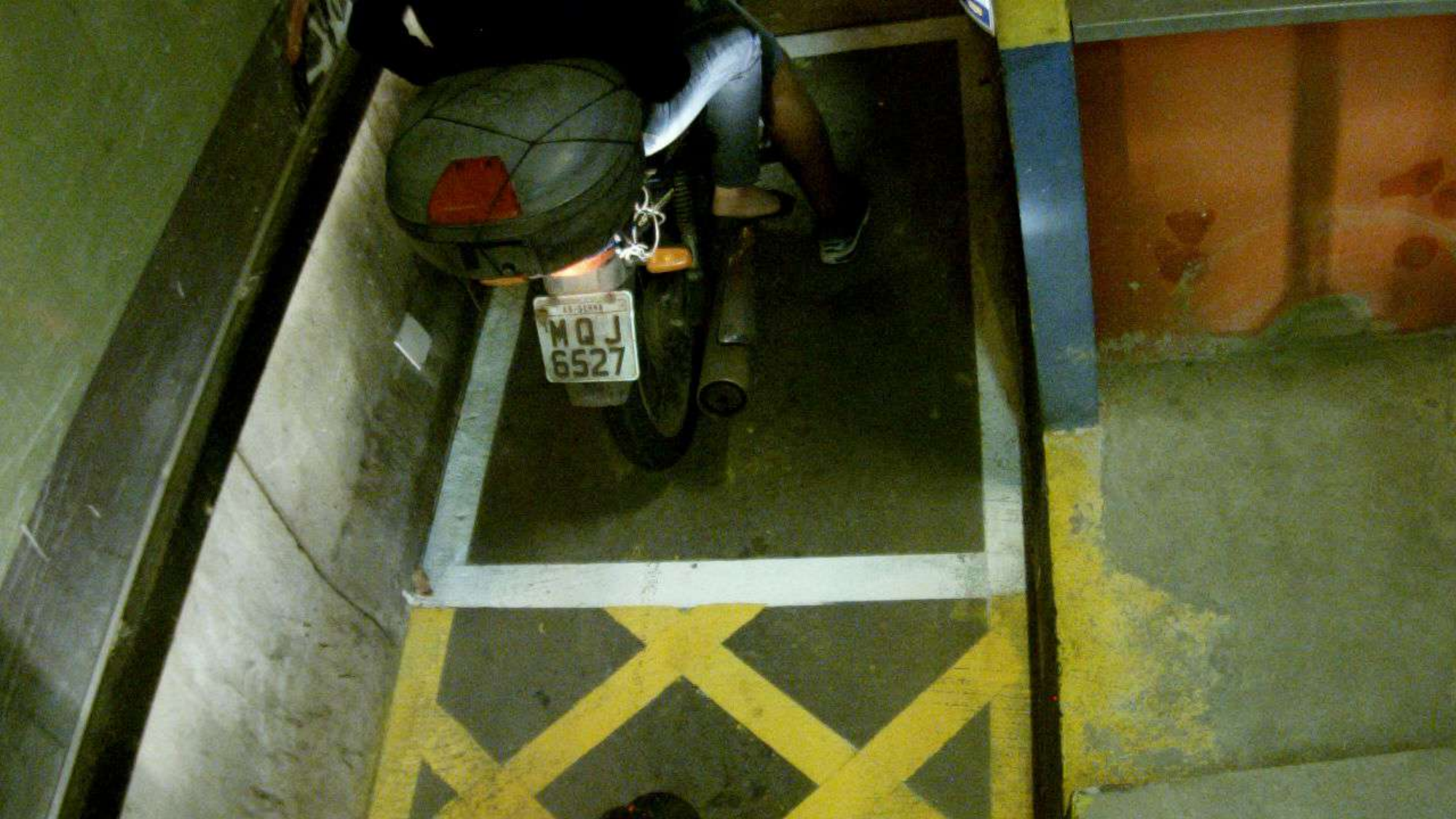}%
\label{fig:motorcycle_2}}

\vspace{-0.25mm}

\caption{Examples of discarded motorcycle images: (a) and (c) were sourced from the RodoSol-ALPR dataset~\cite{laroca2022cross}, while (b) was extracted from the UFPR-ALPR dataset~\cite{laroca2018robust}. Below each image is the corresponding motorcycle's~color. In this figure, the original images were slightly resized for better~viewing.}
\label{fig:motorcycles_rodosol}
\end{figure}

After preprocessing, a total of $28{,}061$ images were obtained.
However, another filtering process was necessary to ensure that every image in the \dataset dataset is suitable for the \gls*{vcr} task.
This involved discarding $10{,}870$ motorcycle images from UFPR-ALPR~($870$) and RodoSol-ALPR~($10{,}000$), as these images represent scenarios where identifying the vehicle color was nearly impossible (see three examples in~\cref{fig:motorcycles_rodosol}).

Another selection process was conducted to remove redundant images from the tracks in the SSIG-SegPlate and UFPR-ALPR datasets.
These datasets contain different vehicle tracks, each comprising a series of sequential frames extracted from videos focusing on a single target vehicle (although other vehicles may appear in the background)~\cite{goncalves2016benchmark,laroca2018robust}.
Due to the low variability between vehicle images cropped from the same track, only one vehicle image per track was selected.
The criterion for selection was to choose the middle image from each track (e.g., if a track contains 30 images, the $15$th image was selected).
As a result, $6{,}278$ vehicle images were excluded, and $10{,}913$ were~retained.

A similar case occurs in the Vehicle-Rear dataset, as it also includes sequential frames extracted from videos.
To address heavily occluded images resulting from overlapping vehicles within the camera's field of view, preference was given to images that clearly depict the vehicle's body.
When multiple images met this criterion, selections were made randomly, leading to the exclusion of $373$ images from the dataset.
Following this process, $10{,}540$ vehicle images were~kept.

In the RodoSol-ALPR dataset, while not consisting of sequential frames extracted from videos, individual vehicles appear multiple times across different days and times.
To ascertain whether images of the same vehicle represent distinct scenarios~\cite{laroca2023do}, images were grouped based on their license plate annotations and manually verified.
During this review, $418$ images were excluded due to minimal perceptible differences in lighting, pose, or other distinguishing characteristics, while $10{,}122$ images were~retained.

\begin{figure}[!htb]
\centering
\captionsetup[subfigure]{captionskip=1.5pt,font=footnotesize}

\resizebox{0.975\linewidth}{!}{
    \subfloat[Multicolored]{\includegraphics[width=0.33\columnwidth,height=0.25\columnwidth]{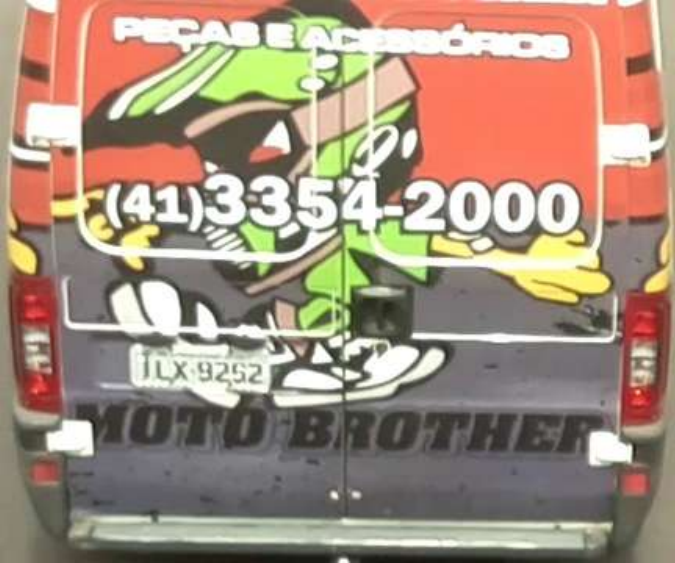}
    \label{fig:removed_fantasy}}
    \subfloat[White]{\includegraphics[width=0.33\columnwidth,height=0.25\columnwidth]{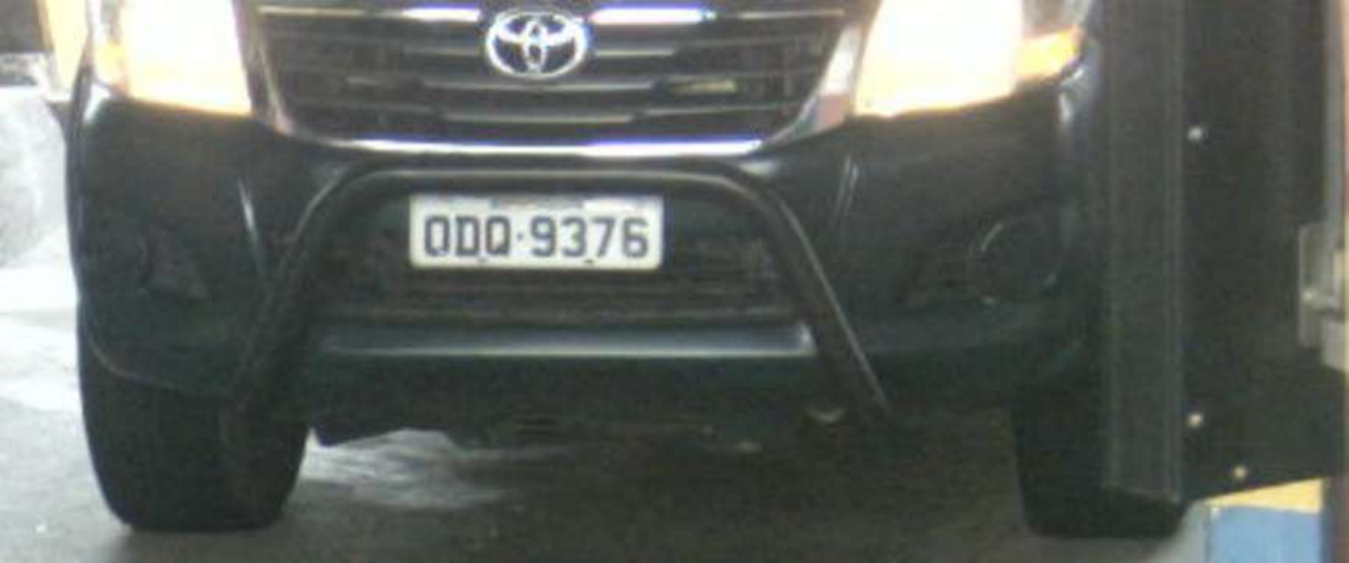}
    \label{fig:removed_occluded_1}}
    \subfloat[White]{\includegraphics[width=0.33\columnwidth,height=0.25\columnwidth]{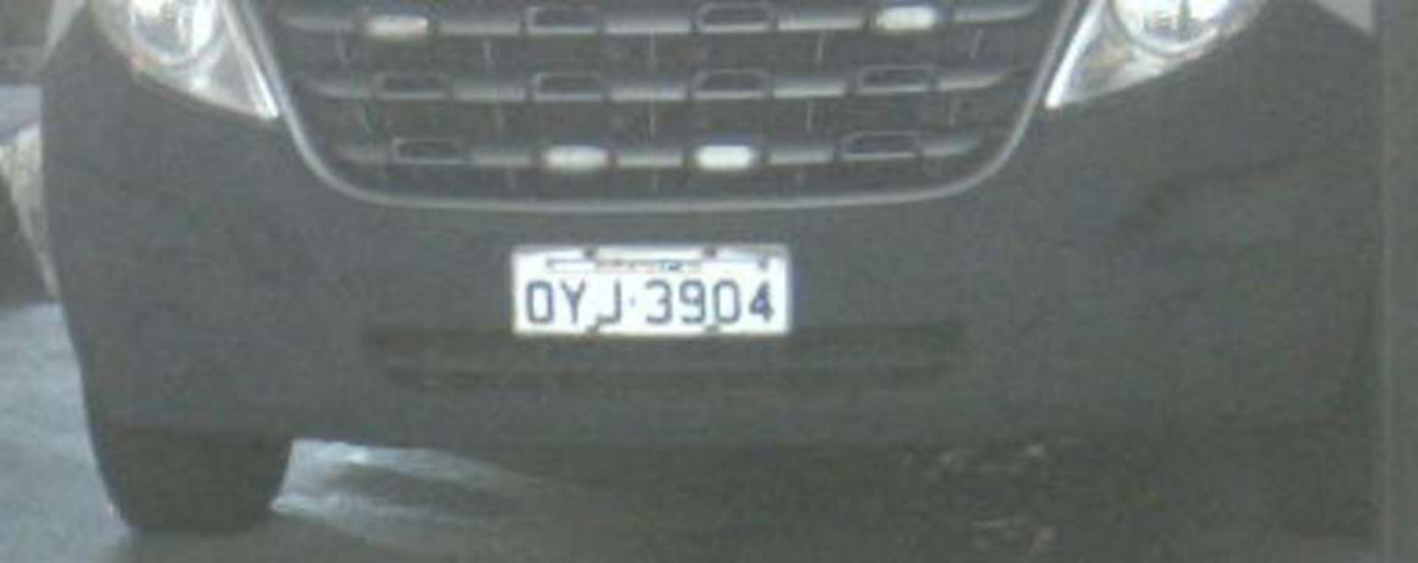}
    \label{fig:removed_occluded_2}}
}

\vspace{-0.25mm}

\caption{Examples of images excluded due to vehicles with multiple colors~(a) and those partially outside the image frame~(b,~c). The colors of the vehicles in~(b) and~(c) cannot be visually determined solely from their front~bumpers. Note that the vehicle images in this figure were resized for improved~visibility.}
\label{fig:ufpr_vc_removed_imgs}
\end{figure}

The final selection process involved removing $83$ images that lacked identifiable colors.
This was mainly observed (i)~when vehicles were heavily occluded or substantially outside the image frame; and (ii)~when vehicles are registered as ``multicolored''\customfootnote{The term ``multicolored'' (a non-literal translation of `\textit{fantasia}' in Portuguese) is used when the vehicle's primary color cannot be determined~\cite{brasil2008transito}.} (a rare occurrence).
\cref{fig:ufpr_vc_removed_imgs} provides examples of images excluded during this selection~process.

\subsection{Annotations}\label{sub_sec:annotations}

As the \dataset dataset derives from established \gls*{alpr} datasets, each image was initially associated with the license plate of the target vehicle.
These annotations enabled the automated retrieval of vehicle information from \gls*{senatran} database, streamlining the annotation process.
In total, $9{,}502$ unique license plates were identified.
However, there were $212$ plates (corresponding to $906$ images) for which vehicle information was unavailable.
These cases required manual annotation.
Finally, vehicles with similar colors were grouped, and the annotations were manually re-validated to ensure the accuracy of each vehicle's~label.

%% file: 4-experiments.tex
\section{Experiments}
\label{sec:experiments}

This section presents a benchmark study conducted on the \dataset dataset.
The study compares four deep learning models on the proposed dataset to evaluate its complexity and identify potential areas for improving \gls*{vcr}. Given the absence of prior studies demonstrating the suitability of these models for \gls*{vcr}, a brief study was performed on the dataset proposed by Chen et al.~\cite{chen2014vehicle}, the results of which are also presented~here.

The rest of this section is structured as follows.
\cref{sub_sec:methodology} covers the experimental methodology.
\cref{sub_sec:results_chen} presents the results obtained on the dataset proposed by Chen et al.~\cite{chen2014vehicle}.
Lastly, \cref{sub_sec:results_ufpr_vc} presents the results achieved on \dataset, highlighting the challenges posed by this~dataset.

\subsection{Methodology}\label{sub_sec:methodology}

The evaluation on the Chen et al.~\cite{chen2014vehicle} dataset and the benchmark on the \dataset dataset share nearly identical methodology.
The only difference is the training protocols applied.
The materials and methods are summarized as~follow:

\subsubsection{Models} 
we explored EfficientNet-V2~\cite{tan2021efficientnetv2}, MobileNet-V3~\cite{howard2019mobilenetv3}, ResNet-34~\cite{he2016residual}, and ViT b16~\cite{dosovitskiy2021vit}.
These models were chosen for their widespread adoption in computer vision tasks and their availability with implementations across various frameworks, which enhances research~reproducibility.

\subsubsection{Dataset splitting} 
the datasets were divided into training, validation, and test subsets using an $8$:$1$:$1$ ratio.
The images were distributed across the different colors for each subset, aiming to mirror the original dataset's class distribution as closely as possible.
When the class distribution did not evenly divide between validation and test subsets, any surplus images were randomly assigned to one of the~subsets.

\subsubsection{Preprocessing}  
all images were resized to $224\times224$ pixels to align with the input size required by the models.
To increase data variability during training, the following transformations were applied to each image in every training~batch:

\begin{itemize}
    \item Affine transformations (with a probability $p$~=~$50$\%), including rotations~($\pm180\degree$), scaling~(from $0.9$ to $1.3$), and shearing~($\pm180\degree$);
    \item Random adjustments to image brightness and contrast~($p$~=~$30$\%), within a limit of $0.2$;
    \item Blur using a generalized normal filter with randomly selected parameters~($p$~=~$40$\%);
    \item A random $72\times72$ pixel section of the image is replaced with random~noise ($p$~=~$25$\%).
\end{itemize}

Finally, every image was normalized using the mean and standard deviation from ImageNet~\cite{deng2009imagenet}.

\subsubsection{Training}
\label{subsub_sec:training}
We employed transfer learning by initializing all models with pre-trained weights from ImageNet~\cite{deng2009imagenet}.
Each network's final fully connected layer was adapted to produce outputs specific to the classes within the dataset being explored.
To be precise, weight adjustments were confined to these layers during the training~phase.

The Adam optimizer~\cite{kingma2015adam} was employed with $\beta_1=0.9$, $\beta_2=0.999$, a batch size of~$128$, weight decay set to~$10$\textsuperscript{-$5$}, and an initial learning rate of~$10$\textsuperscript{-$4$}.
A learning rate reduction scheme was used with a patience value of $10$ and a reduction factor of $10$\textsuperscript{-$1$} upon plateau detection.
Training extended up to a maximum of $400$ epochs, with early stopping configured to halt training if no improvement was observed for $15$ consecutive epochs.
We used the cross-entropy loss~function.

Two different training protocols were implemented for the \dataset dataset: (i)~training with data augmentation only; and (ii)~training with oversampling of minority classes to balance the dataset distribution.
The adopted oversampling method increases the frequency of minority class samples during training by creating synthetic data through data augmentation.
This technique is known to enhance results on imbalanced datasets~\cite{chawla2002smote,shorten2019survey}.
For the dataset proposed by Chen et al.~\cite{chen2014vehicle}, only protocol (i) was adopted, as it proved sufficient to achieve good~results.

\subsubsection{Evaluation}
Each experiment was repeated five times using different dataset splits, and the results are reported based on the average outcomes.
The evaluation metrics used are top-1 accuracy~(Top-1), top-2 accuracy~(Top-2), precision, recall, and F1-score (F1).
These metrics were calculated globally for each iteration using macro averaging.
Running the experiments multiple times helps ensure the reliability of the results, reducing the influence of random variations in the data~splits.

\subsection{Results on the dataset proposed by Chen et al.~\cite{chen2014vehicle}}
\label{sub_sec:results_chen}

This section aims to ``pre-validate'' the models adopted as benchmarks for \dataset, demonstrating their suitability for the \gls*{vcr} task. \cref{tab:results_chen} presents the results obtained for each model on the dataset presented in~\cite{chen2014vehicle}.
Notably, the best-performing model is ViT~b16, achieving an average precision close to those reported in the literature~\cite{chen2014vehicle,hu2015vehicle,fu2018mcff}. Furthermore, the Top-1 and Top-2 accuracies from the evaluated models indicate promising results for the \gls*{vcr}~problem.

\begin{table}[!htp]
    \centering
    \caption{Global metrics~(\%) reached by all models on the dataset proposed by Chen et al.~\cite{chen2014vehicle} (averaged over five runs). The models were trained with the data augmentation~protocol.}
    \label{tab:results_chen}
    \vspace{-2mm}
    \resizebox{0.95\linewidth}{!}{
        \begin{tabular}{lcccccc}
            \toprule
            Model &Top-1 &Top-2 &Precision &Recall &F1 \\\midrule
            EfficientNet-V2~\cite{tan2021efficientnetv2} &$84.6$ &$93.4$ &$84.5$ &$84.6$ &$84.4$ \\
            MobileNet-V3~\cite{howard2019mobilenetv3} &$90.6$ &$96.7$ &$91.7$ &$90.6$ &$91.0$ \\
            ResNet-34~\cite{he2016residual} &$89.0$ &$95.6$ &$91.1$ &$89.0$ &$89.9$ \\
            ViT b16~\cite{dosovitskiy2021vit} &\boldmath$92.8$ &\boldmath$98.0$ &\boldmath$95.3$ &\boldmath$92.8$ &\boldmath$93.9$ \\
            \bottomrule
        \end{tabular}
    }
\end{table}

The results indicate that the explored models are suitable for studying the \gls*{vcr} problem, reinforcing the relevance of the proposed research.
Specifically, with relatively minimal effort, we achieved results comparable to state-of-the-art works on the dataset introduced in~\cite{chen2014vehicle}. 
Hence, we claim that research using this dataset (and others with similar characteristics) does not represent a challenging scenario for \gls*{vcr} evaluation.

\subsection{Results on the \dataset dataset}
\label{sub_sec:results_ufpr_vc}

\cref{tab:results_ufpr_vc} presents the results for each model on the \dataset dataset, considering the two training protocols.
Remarkably, models trained using protocol~(i) achieved better precision and F1 values but exhibited lower accuracy compared to those trained under protocol~(ii).
This discrepancy stems from the higher frequency of minority classes in protocol~(ii).
While this improves overall accuracy, it reduces the precision for the majority classes.
As the test set is also unbalanced, the precision on models trained with protocol (ii) is negatively~affected.

\begin{table}[!ht]
    \centering
    \caption{Global metrics~(\%) reached by all models on the \dataset dataset (averaged over five runs). Protocol (ii) incorporates oversampling of minority classes, whereas (i) does~not.}
    \label{tab:results_ufpr_vc}
    \vspace{-2mm}
    \resizebox{0.95\linewidth}{!}{
        \begin{tabular}{clcccccc}
            \toprule
            Protocol &Model &Top-1 &Top-2 &Precision &Recall &F1 \\
            \midrule
            \multirow{4}{*}{(i)} &EfficientNet-V2~\cite{tan2021efficientnetv2} &$51.2$ &$65.3$ &$65.2$ &$51.2$ &$53.5$ \\
            &MobileNet-V3~\cite{howard2019mobilenetv3} &$50.5$ &$65.4$ &$65.8$ &$50.5$ &$53.1$ \\
             
            &ResNet-34~\cite{he2016residual} &$49.1$ &$60.3$ &$64.3$ &$49.1$ &$52.4$ \\
            &ViT b16~\cite{dosovitskiy2021vit} &\boldmath$59.2$ &\boldmath$71.3$ &\boldmath$76.0$ &\boldmath$59.2$ &\boldmath$62.8$ \\
            \midrule
            \multirow{4}{*}{(ii)} &EfficientNet-V2~\cite{tan2021efficientnetv2} &$55.4$ &$69.5$ &$43.5$ &$55.4$ &$44.6$ \\
            &MobileNet-V3~\cite{howard2019mobilenetv3} &$59.3$ &$73.3$ &$42.6$ &$59.4$ &$45.2$ \\
             
            &ResNet-34~\cite{he2016residual} &$59.3$ &$72.9$ &$47.8$ &$59.3$ &$49.9$ \\
            &ViT b16~\cite{dosovitskiy2021vit} &\boldmath$66.2$ &\boldmath$79.7$ &\boldmath$55.7$ &\boldmath$66.2$ &\boldmath$57.8$ \\
            \bottomrule
        \end{tabular}
    }
\end{table}

\begin{figure}[!htb]
\centering
\includegraphics[width=0.99\linewidth]{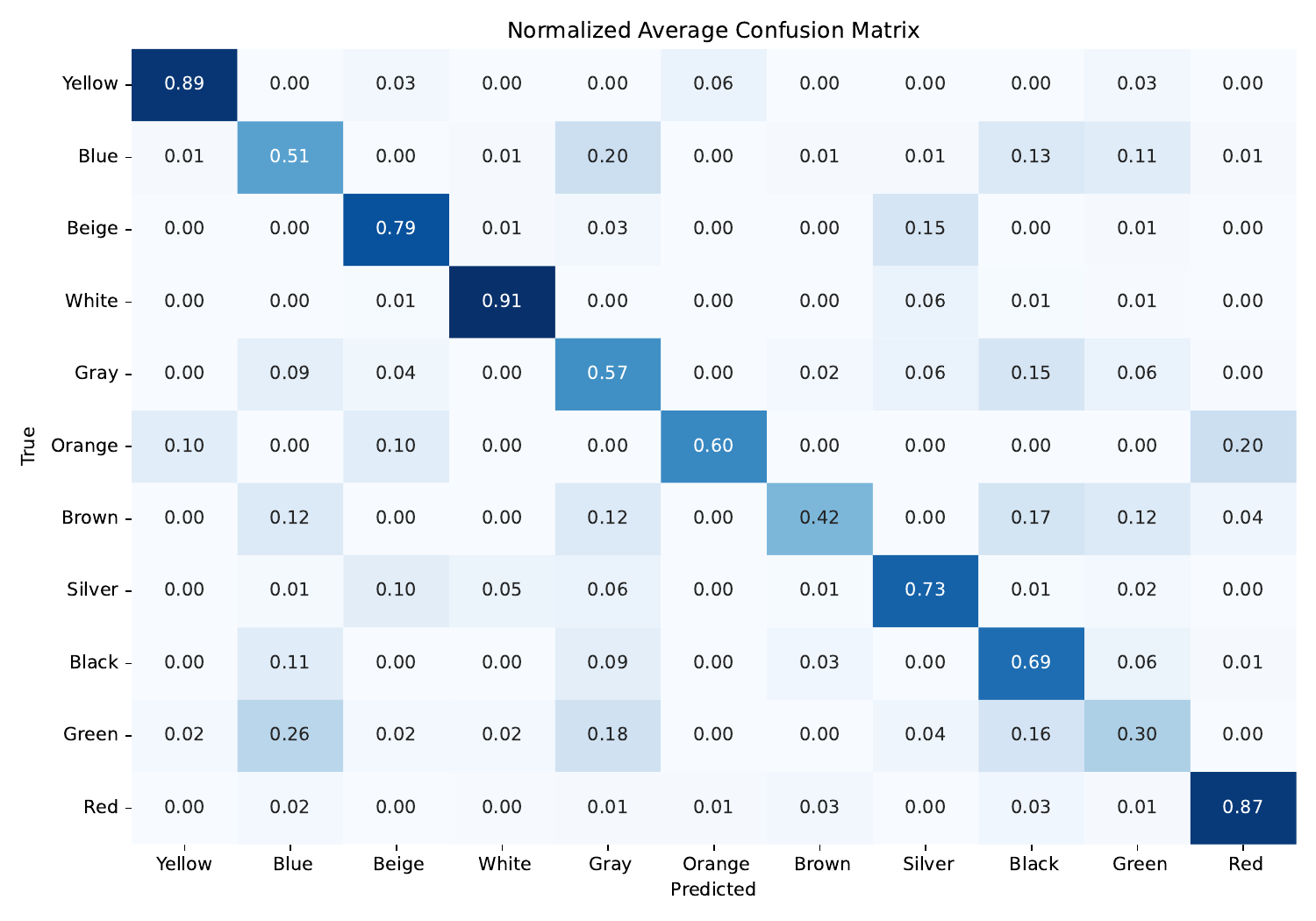}
\vspace{-7.5mm}
\caption{Normalized confusion matrix illustrating the performance of the ViT~b16 model trained with data augmentation and oversampling~techniques.}
\label{fig:ufpr_vc_confusion_matrix_top1}
\end{figure}

Keeping this in mind, we analyzed correct and incorrect predictions using the best model in terms of top-1 accuracy, specifically ViT~b16 trained under protocol~(ii).
\cref{fig:ufpr_vc_confusion_matrix_top1} shows the normalized confusion matrix averaged across the five runs.
It reveals that colors such as yellow, white and red were consistently identified with high accuracy compared to other classes.
Conversely, colors such as brown, blue, green and gray posed challenges for identification.
This difficulty likely stems from dataset characteristics, including significant variations in tone and lighting conditions, potentially causing these colors to be mistaken for shades of gray or~black.

\begin{figure}[!htb]
\centering
\captionsetup[subfigure]{captionskip=1.5pt,labelformat=empty}

\resizebox{0.975\linewidth}{!}{

    \subfloat[\centering \texttt{\phantom{12}\textbf{GT}: White\phantom{i}}\phantom{i}\hspace{\textwidth}\texttt{\textbf{Pred}: Silver}\phantom{i}]{\includegraphics[height=14ex]{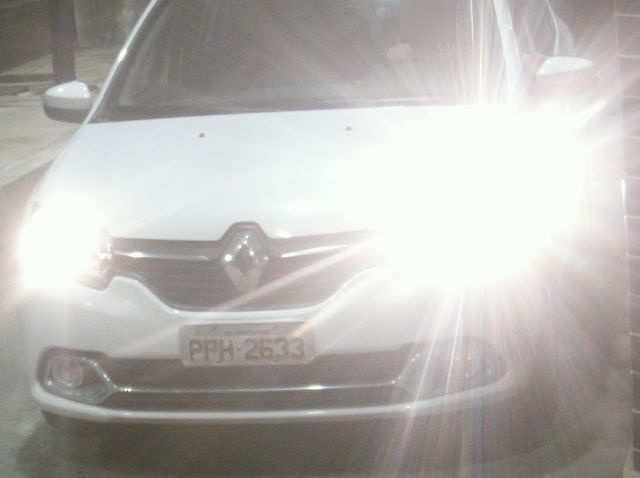}\,
    \label{fig:nighttime_error_white_silver}}
    \subfloat[\centering \texttt{\phantom{12}\textbf{GT}: Red\phantom{12}}\phantom{i}\hspace{\textwidth}\texttt{\textbf{Pred}: White}\phantom{i}]{\includegraphics[height=14ex]{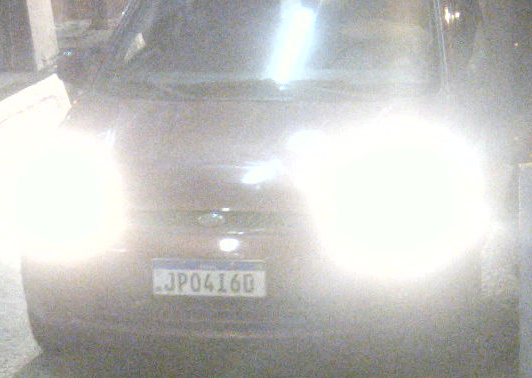}\,
    \label{fig:nighttime_error_red_gray}}
    \subfloat[\centering \texttt{\phantom{12}\textbf{GT}: Black}\phantom{i}\hspace{\textwidth}\texttt{\textbf{Pred}: Gray\phantom{1}}\phantom{i}]{\includegraphics[height=14ex]{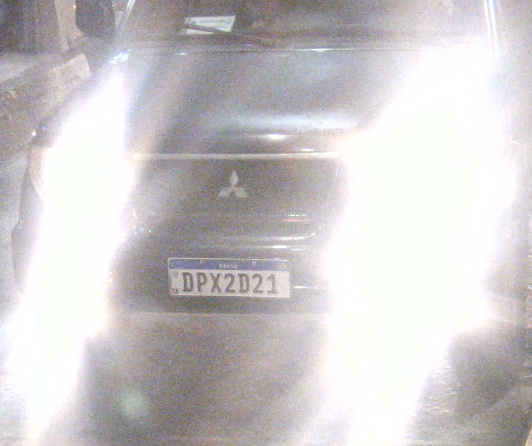}
    \label{fig:nighttime_error_blue_red}}
}

\vspace{-0.25mm}

\caption{Examples of nighttime images that were misclassified.}
\label{fig:nighttime}
\end{figure}

Our analysis uncovered an interesting trend: nighttime images were misclassified at a much higher rate than daytime images.
Specifically, $72$ out of $222$ misclassified images ($32.4$\%) were captured at night.
Nighttime images likely constitute less than $10$\% of the \dataset dataset, although the exact percentage is unknown as capture times are not labeled in the original datasets.
This discrepancy is likely due to the inherent challenges of nighttime scenes, such as high illumination and overexposure from vehicle headlights (see \cref{fig:nighttime}).
The causes behind the remaining misclassifications were less~apparent.

Finally, we analyzed the model's second-choice predictions~(top $2$) for images it initially misclassified~(errors in terms of top-$1$ accuracy).
Notably, colors like beige, white, gray, silver, and black achieved over $50$\% classification accuracy in these cases.
In other words, the model's second prediction was correct for more than half of the misclassified images in these color categories.
However, it is important to note that an average of $44.4$\% of the nighttime images remained incorrectly classified even considering the top-$2$~predictions.

%% file: 5-conclusions.tex
\section{Conclusions}
\label{sec:conclusions}

\glsreset{vcr} %

This study revealed shortcomings in existing \gls{vcr} datasets, emphasizing their inadequacy in replicating real-world, unconstrained scenarios.
To address this issue, we compiled the \dataset dataset. 
It comprises $10{,}039$ images featuring adverse scenarios, such as various viewpoints, uneven lighting, and nighttime scenes, across $11$ color classes.
\ifieee
Our study using four deep learning models demonstrated that \dataset presents significant challenges for \gls*{vcr}, particularly in nighttime scenarios, which accounted for $\approx33$\% of the errors by the best-performing model despite representing a much smaller portion of the~dataset.
\else
A benchmark study using four deep learning models demonstrated that \dataset presents significant challenges for \gls*{vcr}, particularly in nighttime scenarios, which accounted for $\approx33$\% of the errors by the best-performing model despite representing a much smaller portion of the~dataset.
\fi

This study identifies remaining scenarios in \gls*{vcr} that require further investigation.
Developing novel methods for robust color recognition under adverse conditions is essential to improve the reliability of these approaches in real-world applications.
We hope this work serves as a catalyst for \gls*{vcr} research in adverse conditions, encouraging future studies to address progressively more difficult scenarios.

An important future research direction is tackling the challenges of \gls*{vcr} in nighttime scenes.
This endeavor would likely involve investigating advanced preprocessing methods and designing specialized architectures to enhance current results.
Additionally, we plan to enrich the dataset by incorporating more vehicle attributes, such as type (e.g., sedan, hatchback, truck), make, and model.
This would enable the integration of color recognition with fine-grained vehicle classification tasks, potentially through a multi-task learning~framework.

%% file: 0-acknowledgments.tex
\section*{\uppercase{Acknowledgments}}

\iffinal
    This study was financed in part by the \textit{Coordenação de Aperfeiçoamento de Pessoal de Nível Superior - Brasil~(CAPES)} - Finance Code 001, and in part by the \textit{Conselho Nacional de Desenvolvimento Científico e Tecnológico~(CNPq)} (\#~315409/2023-1).
    \ifieee
        We thank NVIDIA for donating the Quadro RTX $8000$ GPU used for this research.
    \else
        We thank the support of NVIDIA Corporation with the donation of the Quadro RTX $8000$ GPU used for this research.
    \fi
\fi